\documentclass[11pt,a4paper]{article}
\pdfoutput=1
\usepackage[hyperref]{acl2020}
\usepackage{times}
\usepackage{latexsym}

\newcommand{\egcvalue}[1]{\textbf{\textit{#1}}}
\usepackage{enumitem}
\usepackage{color}
\usepackage{tcolorbox}
\usepackage{tikz}
\usepackage{amssymb}

\usepackage{microtype}

\aclfinalcopy 

\definecolor{azure}{rgb}{0.0, 0.5, 1.0}
\tcbuselibrary{skins}
\tcbset{enhanced}
\newcommand{\qsecbox}[1]{\begin{tcolorbox}[left=1mm,right=1mm,boxrule=0.2mm,leftrule=2mm,drop fuzzy shadow,colframe=lightgray,frame style={left color=azure!90!lightgray}]#1\end{tcolorbox}}

\makeatletter
\newcommand\footnoteref[1]{\protected@xdef\@thefnmark{\ref{#1}}\@footnotemark}
\makeatother

\title{The Human Evaluation Datasheet 1.0: A Template for Recording\\Details of Human Evaluation Experiments in NLP}

\author{
  Anastasia Shimorina \\
  Universit\'e de Lorraine / LORIA\\
  F-54000 Nancy, France \\
  \texttt{anastasia.shimorina@loria.fr} \And
  Anya Belz \\
  University of Brighton\\
  Lewes Road, Brighton, UK \\
  \texttt{a.s.belz@brighton.ac.uk}
}

\date{}

\begin{document}
\maketitle
\begin{abstract}
This paper introduces the Human Evaluation Datasheet, a template for recording the details of individual human evaluation experiments in Natural Language Processing (NLP). 
Originally taking inspiration from seminal papers by \citet{bender-friedman-2018-data},  \citet{mitchell2019modelcards}, and \citet{gebru2018datasheets}, the Human Evaluation Datasheet is intended to facilitate the recording of properties of human evaluations in sufficient detail, and with sufficient standardisation, to support comparability, meta-evaluation, and reproducibility tests. 
\end{abstract}

\section{Introduction}

The Human Evaluation Datasheet (HEDS) is a template for describing single human evaluation experiments (see Section~\ref{sec:kind} for explanation of this term). It uses multiple-choice questions where possible, for ease of comparability across experiments.
The HEDS template is intended to be generally applicable to human evaluations across NLP. Some of the questions in Version 1.0 are geared towards systems that produce language as output, but have `Other' options for entering alternative information. The plan is to generalise this and other aspects of the template in future versions of HEDS. 

Questions 2.1--2.5 relating to evaluated system, and 4.3.1--4.3.8 relating to response elicitation, are based on \citet{howcroft-etal-2020-twenty}, with some significant changes. Questions 4.1.1--4.2.3 relating to quality criteria, and some of the questions about system outputs, evaluators, and experimental design (3.1.1--3.2.3, 4.3.5, 4.3.6, 4.3.9--4.3.11) are based on \citet{belz-etal-2020-disentangling}. 
HEDS was also informed by \citet{van2019best, vanderlee2021} and by  \citet{gehrmann2021gem}'s\footnote{\footnotesize \url{https://gem-benchmark.com/data\_cards/guide}} data card guide.

More generally, the original inspiration for creating a `datasheet' for describing human evaluation experiments of course comes from seminal papers by \citet{bender-friedman-2018-data}, \citet{mitchell2019modelcards} and \citet{gebru2018datasheets}.

We envisage this datasheet to undergo changes in the future, as more teams of authors complete it and we discover more ways of improving it. Changes will be indicated in the title by increments to the version number, and a change log in the body of the paper.
In its template form as well as in this paper, HEDS is divided into five sections:
\begin{enumerate}\itemsep=0pt
    \item Paper and Resources (Questions 1.1--1.3, covered in Section~\ref{sec:paper-resources} of this paper);
    \item Evaluated System (Questions 2.1--2.5, Section~\ref{sec:system} below);
    \item Output Sample, Evaluators and Experimental Design (Questions 3.1.1--3.3.8, Section~\ref{sec:design} below);
    \item Quality Criteria (Questions 4.1.1--4.3.11, Section~\ref{sec:criteria} below); 
    \item Ethics (Questions 5.1--5.4, Section~\ref{sec:ethics} below).
\end{enumerate}

\section{Package Resources}\label{sec:resources}

The Human Evaluation Datasheet 1.0 package (HEDS 1.0) consists of the following three resources:

\begin{enumerate}\itemsep=0pt
    \item The HEDS 1.0 form: available at \url{https://forms.gle/MgWiKVu7i5UHeMNQ9};
    \item Description and completion guidance: this document;
    \item Scripts for automatically converting between the HEDS 1.0 online form and alternative Markdown and LaTeX template formats: available (from May 2021) at \url{https://github.com/Shimorina/human-evaluation-datasheet}.
\end{enumerate}

\noindent A repository of completed HEDS datasheets is also planned.

\section{What kind of evaluation experiment can this sheet be used for?}\label{sec:kind}

Our aim is to make this sheet suitable for all \emph{single} human evaluation experiments in NLP. Human evaluations in NLP typically get participants to assess system outputs, but the HEDS sheet also accommodates what we call `human-authored stand-ins' below, i.e.\ manually created `system outputs' (e.g.\ in a wizard-of-oz scenario or when reference outputs are included in an evaluation) evaluated in a way that can at least in principle be used to evaluate actual system outputs.

By `single human evaluation experiment' we mean one that evaluates a single set of directly comparable systems in a single experimental design, but may assess multiple quality criteria. We refer to the single human evaluation experiment that the template is being completed for simply as `the evaluation experiment' in questions below.

It should soon become apparent if you're trying to complete the sheet for something that isn't a single evaluation experiment, because you will end up wanting to put `in one part of the experiment we did A, in another we did B', for the experimental design and other questions in Part~3.

\section{Paper and Supplementary Resources (Questions 1.1--1.3)}\label{sec:paper-resources}

Questions 1.1--1.3 record bibliographic and related information. These are straightforward and don't warrant much in-depth explanation.

\vspace{-.3cm}
\subsection*{\qsecbox{Question 1.1: Link to paper reporting the evaluation experiment. If the paper reports more than one experiment, state which experiment you're completing this sheet for. Or, if applicable, enter `for preregistration.'}}

\noindent\textit{What to enter in the text box}: a link to an online copy of the main reference for the human evaluation experiment, identifying which of the experiments the form is being completed for if there are several. If the experiment hasn't been run yet, and the form is being completed for the purpose of submitting it for preregistration, simply enter `for preregistration'.

\vspace{-.3cm}
\subsection*{\qsecbox{Question 1.2: Link to website providing resources used in the evaluation experiment (e.g.\ system outputs, evaluation tools, etc.). If there isn't one, enter `N/A'.}}

\noindent\textit{What to enter in the text box}: link(s) to any resources used in the evaluation experiment, such as system outputs, evaluation tools, etc.\ If there aren't any publicly shared resources (yet), enter `N/A’.

\vspace{-.3cm}
\subsection*{\qsecbox{Question 1.3: Name, affiliation and email address of person completing this sheet, and of contact author if different.}}

\noindent\textit{What to enter in the text box}: names, affiliations and email addresses as appropriate.

\section{System (Questions 2.1--2.5)}\label{sec:system}

Questions 2.1--2.5 record information about the system(s) (or human-authored stand-ins) whose outputs are evaluated in the Evaluation experiment that this sheet is being completed for.

The input, output, and task questions in this section are closely interrelated: the value for one partially determines the others, as indicated for some combinations in Question 2.3.

\vspace{-.3cm}
\subsection*{\qsecbox{Question 2.1: What type of input do the evaluated system(s) take? Select all that apply. If none match, select `Other' and describe.}}\label{sec:input}
\vspace{-.1cm}

Describe the type of input, where input refers to the representations and/or data structures shared by all evaluated systems. 

This question is about input type, regardless of number. E.g.\ if the input is a set of documents, you would still select \textit{text: document} below. 

\vspace{.3cm}
\noindent\textit{Check-box options (select all that apply)}: 
\vspace{-.1cm}

\begin{enumerate}[itemsep=0cm,leftmargin=0.5cm,label={\small $\square$}]
            \item \egcvalue{raw/structured data}: numerical, symbolic, and other data, possibly structured into trees, graphs, graphical models, etc. May be the input e.g.\ to Referring Expression Generation (REG), end-to-end text generation, etc. {NB}: excludes linguistic structures.
            
            \item \egcvalue{deep linguistic representation (DLR)}: any of a variety of deep, underspecified, semantic representations, such as abstract meaning representations \citep[AMRs;][]{banarescu-etal-2013-abstract} or discourse representation structures \citep[DRSs;][]{kamp-reyle2013discourse}.
            
            \item \egcvalue{shallow linguistic representation (SLR)}: any of a variety of shallow, syntactic representations, e.g.\ Universal Dependency (UD) structures; typically the input to surface realisation.
            
            \item \egcvalue{text: subsentential unit of text}: a unit of text shorter than a sentence, e.g.\ Referring Expressions (REs), verb phrase, text fragment of any length; includes titles/headlines.
            
            \item \egcvalue{text: sentence}: a single sentence (or set of sentences).
            
            \item \egcvalue{text: multiple sentences}: a sequence of multiple sentences, without any document structure (or a set of such sequences). 
            
            \item \egcvalue{text: document}: a text with document structure, such as a title, paragraph breaks or sections, e.g.\ a set of news reports for summarisation.
            
            \item \egcvalue{text: dialogue}: a dialogue of any length, excluding a single turn which would come under one of the other text types.
            
            \item \egcvalue{text: other}: input is text but doesn't match any of the above \textit{text:*} categories.
            
            \item \egcvalue{speech}: a recording of speech.
            
            \item \egcvalue{visual}: an image or video.
            
            \item \egcvalue{multi-modal}: catch-all value for any combination of data and/or linguistic representation and/or visual data etc.
            
            \item \egcvalue{control feature}: a feature or parameter specifically present to control a property of the output text, e.g.\ positive stance, formality, author style.
            
            \item \egcvalue{no input (human generation)}: human generation\footnote{\label{human-generation}We use the term `human generation' where the items being evaluated have been created manually, rather than generated by an automatic system.}, therefore no system inputs.
            
            \item \egcvalue{other (please specify)}: if input is none of the above, choose this option and describe it. 
            
\end{enumerate}

\vspace{-.3cm}
\subsection*{\qsecbox{Question 2.2: What type of output do the evaluated system(s) generate? Select all that apply. If none match, select `Other' and describe.}}\label{sec:output}

Describe the type of input, where input refers to the representations and/or data structures shared by all evaluated systems. 

This question is about input type, regardless of number. E.g.\ if the output is a set of documents, you would still select \textit{text: document} below. 

Note that the options for outputs are the same as for inputs minus the \textit{control feature} option.

\vspace{.3cm}
\noindent\textit{Check-box options (select all that apply)}: 
\vspace{-.1cm}

\begin{enumerate}[itemsep=0cm,leftmargin=0.5cm,label={\small $\square$}]
            \item \egcvalue{raw/structured data}: numerical, symbolic, and other data, possibly structured into trees, graphs, graphical models, etc. May be the input e.g.\ to Referring Expression Generation (REG), end-to-end text generation, etc. {NB}: excludes linguistic structures.
            
            \item \egcvalue{deep linguistic representation (DLR)}: any of a variety of deep, underspecified, semantic representations, such as abstract meaning representations \citep[AMRs;][]{banarescu-etal-2013-abstract} or discourse representation structures \citep[DRSs;][]{kamp-reyle2013discourse}.
            
            \item \egcvalue{shallow linguistic representation (SLR)}: any of a variety of shallow, syntactic representations, e.g.\ Universal Dependency (UD) structures; typically the input to surface realisation.
            
            \item \egcvalue{text: subsentential unit of text}: a unit of text shorter than a sentence, e.g.\ Referring Expressions (REs), verb phrase, text fragment of any length; includes titles/headlines.
            
            \item \egcvalue{text: sentence}: a single sentence (or set of sentences).
            
            \item \egcvalue{text: multiple sentences}: a sequence of multiple sentences, without any document structure (or a set of such sequences). 
            
            \item \egcvalue{text: document}: a text with document structure, such as a title, paragraph breaks or sections, e.g.\ a set of news reports for summarisation.
            
            \item \egcvalue{text: dialogue}: a dialogue of any length, excluding a single turn which would come under one of the other text types.
            
           \item \egcvalue{text: other}: select if output is text but doesn't match any of the above \textit{text:*} categories.
            
            \item \egcvalue{speech}: a recording of speech.
            
            \item \egcvalue{visual}: an image or video.
            
            \item \egcvalue{multi-modal}: catch-all value for any combination of data and/or linguistic representation and/or visual data etc.
            
            \item \egcvalue{human-generated `outputs'}: manually created stand-ins exemplifying outputs.\footnoteref{human-generation}
            
            \item \egcvalue{other (please specify)}: if output is none of the above, choose this option and describe it. 
            
        \end{enumerate}

\vspace{-.3cm}
\subsection*{\qsecbox{Question 2.3: How would you describe the task that the evaluated system(s) perform in mapping the inputs in Q2.1 to the outputs in Q2.2? Occasionally, more than one of the options below may apply. If none match, select `Other' and describe.}}\label{sec:task}
\vspace{-.1cm}

This field records the task performed by the system(s) being evaluated. This is independent of the application domain (financial reporting, weather forecasting, etc.), or the specific method (rule-based, neural, etc.) implemented in the system. We indicate mutual constraints between inputs, outputs and task for some of the options below.

\vspace{.3cm}
\noindent\textit{Check-box options (select all that apply)}:  
\vspace{-.1cm}

\begin{enumerate}[itemsep=0cm,leftmargin=0.5cm,label={\small $\square$}]
    \item \egcvalue{content selection/determination}: selecting  the specific content that will be expressed in the generated text from a representation of possible content. This could be attribute selection for REG (without the surface realisation step). Note that the output here is not text.
    
    \item \egcvalue{content ordering/structuring}: assigning an order and/or structure to content to be included in generated text. Note that the output here is not text.
    
    \item \egcvalue{aggregation}: converting inputs (typically \textit{deep linguistic representations} or \textit{shallow linguistic representations}) in some way in order to reduce redundancy (e.g.\  representations for `they like swimming', `they like running' $\rightarrow$ representation for `they like swimming and running').
   
    \item \egcvalue{referring expression generation}: generating \textit{text} to refer to a given referent, typically represented in the input as a set of attributes or a linguistic representation. 
    
    \item \egcvalue{lexicalisation}: associating (parts of) an input representation with specific lexical items to be used in their realisation. 
    
    \item \egcvalue{deep generation}: one-step text generation from \textit{raw/structured data} or \textit{deep linguistic representations}. One-step means that no intermediate representations are passed from one independently run module to another.
    
    \item \egcvalue{surface realisation (SLR to text)}: one-step text generation from \textit{shallow linguistic representations}. One-step means that no intermediate representations are passed from one independently run module to another.
    
    \item \egcvalue{feature-controlled text generation}: generation of text that varies along specific dimensions where the variation is controlled via \textit{control feature}s specified as part of the input. Input is a non-textual representation (for feature-controlled text-to-text generation select the matching text-to-text task). 
    
    \item \egcvalue{data-to-text generation}: generation from \textit{raw/structured data} which may or may not include some amount of content selection as part of the generation process. Output is likely to be \textit{text:*} or \textit{multi-modal}.
    
    \item \egcvalue{dialogue turn generation}: generating a dialogue turn (can be a greeting or closing) from a representation of dialogue state and/or last turn(s), etc. 

    \item \egcvalue{question generation}: generation of questions from given input text and/or knowledge base such that the question can be answered from the input.
    
    \item \egcvalue{question answering}: input is a question plus optionally a set of reference texts and/or knowledge base, and the output is the answer to the question.
    
    \item \egcvalue{paraphrasing/lossless simplification}: text-to-text generation where the aim is to preserve the meaning of the input while changing its wording. This can include the aim of changing the text on a given dimension, e.g.\ making it simpler, changing its stance or sentiment, etc., which may be controllable via input features. Note that this task type includes meaning-preserving text simplification (non-meaning preserving simplification comes under \textit{compression/lossy simplification} below).
    
    \item \egcvalue{compression/lossy simplification}: text-to-text generation that has the aim to generate a shorter, or shorter and simpler, version of the input text. This will normally affect meaning to some extent, but as a side effect, rather than the primary aim, as is the case in \textit{summarisation}.
    
    \item \egcvalue{machine translation}: translating text in a source language to text in a target language while maximally preserving the meaning. 
    
    \item \egcvalue{summarisation (text-to-text)}: output is an extractive or abstractive summary of the important/relevant/salient content of the input  document(s).

    \item \egcvalue{end-to-end text generation}: use this option if the single system task corresponds to more than one of tasks above, implemented either as separate modules pipelined together, or as one-step generation, other than \textit{deep generation} and \textit{surface realisation}.
    
    \item \egcvalue{image/video description}: input includes \textit{visual}, and the output describes it in some way.
    
    \item \egcvalue{post-editing/correction}: system edits and/or corrects the input text (typically itself the textual output from another system) to yield an improved version of the text.
       
    \item \egcvalue{other (please specify)}: if task is none of the above, choose this option and describe it.
    \end{enumerate}

\vspace{-.3cm}
\subsection*{\qsecbox{Question 2.4: Input Language(s), or `N/A'.}}
\vspace{-.1cm}

This field records the language(s) of the inputs accepted by the system(s) being evaluated.
        
\vspace{.3cm}
\noindent\textit{What to enter in the text box}: any  language name(s) that apply, mapped to standardised full language names in ISO 639-1\footnote{\label{iso}\url{https://en.wikipedia.org/wiki/List_of_ISO_639-1_codes}}. E.g.\ English, Herero, Hindi. 
If no language is accepted as (part of) the input, enter `N/A'.
        
\vspace{-.3cm}
\subsection*{\qsecbox{Question 2.5: Output Language(s), or `N/A'.}}
\vspace{-.1cm}

This field records the language(s) of the outputs generated by the system(s) being evaluated.
        
\vspace{.2cm}
\noindent\textit{What to enter in the text box}: any  language name(s) that apply, mapped to standardised full language names in ISO 639-1 (2019)\footnoteref{iso}. E.g.\ English, Herero, Hindi. 
If no language is generated, enter `N/A'.

\section{Output Sample, Evaluators, Experimental Design}\label{sec:design}

\subsection{Sample of system outputs (or human-authored stand-ins) evaluated (Questions 3.1.1--3.1.3)}

Questions 3.1.1--3.1.3 record information about the size of the sample of outputs (or human-authored stand-ins) evaluated per system, how the sample was selected, and what its statistical power is.

\vspace{-.3cm}
\subsection*{\qsecbox{Question 3.1.1: How many system outputs (or other evaluation items) are evaluated per system in the evaluation experiment? Answer should be an integer.}}

\noindent\textit{What to enter in the text box}: The number of system outputs (or other evaluation items) that are evaluated per system by at least one evaluator in the experiment, as an integer.

\vspace{-.3cm}
\subsection*{\qsecbox{Question 3.1.2: How are system outputs (or other evaluation items) selected for inclusion in the evaluation experiment? If none match, select `Other' and describe.}}

\noindent\textit{Multiple-choice options (select one)}:  
\vspace{-.1cm}

\begin{enumerate}[itemsep=0cm,leftmargin=0.5cm,label={\LARGE $\circ$}]
    \item \egcvalue{by an automatic random process from a larger set}: outputs were selected for inclusion in the experiment by a script using a pseudo-random number generator; don't use this option if the script selects every $n$th output (which is not random). 
    \item \egcvalue{by an automatic random process but using stratified sampling over given properties}: use this option if selection was by a random script as above, but with added constraints ensuring that the sample is representative of the set of outputs it was selected from, in terms of given properties, such as sentence length, positive/negative stance, etc.
    \item \egcvalue{by manual, arbitrary selection}: output sample was selected by hand, or automatically from a manually compiled list, without a specific selection criterion.
    \item \egcvalue{by manual selection aimed at achieving balance or variety relative to given properties}: selection by hand as above, but with specific selection criteria, e.g.\ same number of outputs from each time period.
    \item \egcvalue{Other (please specify)}: if selection method is none of the above, choose this option and describe it.
\end{enumerate}

\vspace{-.3cm}
\subsection*{\qsecbox{Question 3.1.3: What is the statistical power of the sample size?}}

\noindent\textit{What to enter in the text box}:  The results of a statistical power calculation on the output sample: provide numerical results and a link to the script used (or another way of identifying the script). See, e.g., \citet{card-etal-2020-little}.

\subsection{Evaluators (Questions 3.2.1--3.2.4)}

Questions 3.2.1--3.2.4 record information about the evaluators participating in the experiment.

\vspace{-.3cm}
\subsection*{\qsecbox{Question 3.2.1:  How many evaluators are there in this experiment? Answer should be an integer.}}

\noindent\textit{What to enter in the text box}: the total number of evaluators participating in the experiment, as an integer.

\vspace{-.3cm}
\subsection*{\qsecbox{Question 3.2.2:  What kind of evaluators are in this experiment? Select all that apply. If none match, select `Other' and describe. In all cases, provide details in the text box under `Other'.}}

\noindent\textit{Check-box options (select all that apply)}:  
\vspace{-.1cm}

\begin{enumerate}[itemsep=0cm,leftmargin=0.5cm,label={\small $\square$}]
    \item \egcvalue{experts}: participants are considered domain experts, e.g.\ meteorologists evaluating a weather forecast generator, or nurses evaluating an ICU report generator.
    \item \egcvalue{non-experts}: participants are not domain experts.
    \item \egcvalue{paid (including non-monetary compensation such as course credits)}: participants were given some form of compensation for their participation, including vouchers, course credits, and reimbursement for travel unless based on receipts.
    \item \egcvalue{not paid}: participants were not given compensation of any kind.
    \item \egcvalue{previously known to authors}: (one of the) researchers running the experiment knew some or all of the participants before recruiting them for the experiment.
    \item \egcvalue{not previously known to authors}: none of the researchers running the experiment knew any of the participants before recruiting them for the experiment.
    \item \egcvalue{evaluators include one or more of the authors}: one or more researchers running the experiment was among the participants.
    \item \egcvalue{evaluators do not include any of the authors}: none of the researchers running the experiment were among the participants.
    \item \egcvalue{Other} (fewer than 4 of the above apply): we believe you should be able to tick 4 options of the above. If that's not the case, use this box to explain.
\end{enumerate}

\vspace{-.3cm}
\subsection*{\qsecbox{Question 3.2.3:  How are evaluators recruited?}}

\noindent\textit{What to enter in the text box}: Please explain how your evaluators are recruited. Do you send emails to a given list? Do you post invitations on social media? Posters on university walls? Were there any gatekeepers involved? What are the exclusion/inclusion criteria? 

\vspace{-.3cm}
\subsection*{\qsecbox{Question 3.2.4: What training and/or practice are evaluators given before starting on the evaluation itself?}}

\noindent\textit{What to enter in the text box}: Use this space to describe any training evaluators were given as part of the experiment to prepare them for the evaluation task, including any practice evaluations they did. This includes any introductory explanations they're given, e.g.\ on the start page of an online evaluation tool.

\vspace{-.3cm}
\subsection*{\qsecbox{Question 3.2.5:  What other characteristics do the evaluators have, known either because these were qualifying criteria, or from information gathered as part of the evaluation?}}

\noindent\textit{What to enter in the text box}: Use this space to list any characteristics not covered in previous questions that the evaluators are known to have, either because evaluators were selected on the basis of a characteristic, or because information about a characteristic was collected as part of the evaluation. This might include geographic location of IP address, educational level, or demographic information such as gender, age, etc. Where characteristics differ among evaluators (e.g.\ gender, age, location etc.), also give numbers for each subgroup.

\subsection{Experimental design (Questions 3.3.1--3.3.8)}

Questions~3.3.1--3.3.8 record information about the experimental design of the evaluation experiment.

\vspace{-.3cm}
\subsection*{\qsecbox{Question 3.3.1:  Has the experimental design been preregistered? If yes, on which registry?}}

\noindent\textit{What to enter in the text box}: State `Yes' or `No'; if `Yes' also give the name of the registry and a link to the registration page for the experiment.

\subsection*{\qsecbox{Question 3.3.2: How are responses collected? E.g.\ paper forms, online survey tool, etc.}}

\noindent\textit{What to enter in the text box}: Use this space to describe how you collected responses, e.g.\ paper forms, Google forms, SurveyMonkey, Mechanical Turk, CrowdFlower, audio/video recording, etc. 

\subsection*{\qsecbox{Question 3.3.3:  What quality assurance methods are used? Select all that apply.   If none match, select `Other' and describe. In all cases, provide details in the text box under `Other'.}}
\vspace{-.1cm}

\vspace{.3cm}
\noindent\textit{Check-box options (select all that apply)}:  
\vspace{-.1cm}

\begin{enumerate}[itemsep=0cm,leftmargin=0.5cm,label={\small $\square$}]
    \item \egcvalue{evaluators are required to be native speakers of the language they evaluate}: mechanisms are in place to ensure all participants are native speakers of the language they evaluate.
    \item \egcvalue{automatic quality checking methods are used during/post evaluation}: evaluations are checked for quality by automatic scripts during or after evaluations, e.g.\ evaluators are given known bad/good outputs to check they're given bad/good scores on MTurk.
    \item \egcvalue{manual quality checking methods are used during/post evaluation}: evaluations are checked for quality by a manual process  during or after evaluations, e.g.\ scores assigned by evaluators are monitored by researchers conducting the experiment.
    \item \egcvalue{evaluators are excluded if they fail quality checks (often or badly enough)}: there are conditions under which evaluations produced by participants are not included in the final results due to quality issues.
    \item \egcvalue{some evaluations are excluded because of failed quality checks}: there are conditions under which some (but not all) of the evaluations produced by some participants are not included in the final results due to quality issues.
    \item \egcvalue{none of the above}: tick this box if none of the above apply.
    \item \egcvalue{Other (please specify)}: use this box to describe any other quality assurance methods used during or after evaluations, and to provide additional details for any of the options selected above.
\end{enumerate}

\subsection*{\qsecbox{Question 3.3.4:  What do evaluators see when carrying out evaluations? Link to screenshot(s) and/or describe the evaluation interface(s).}}
\vspace{-.1cm}

\vspace{.3cm}
\noindent\textit{What to enter in the text box}: Use this space to describe the interface, paper form, etc.\ that evaluators see when they carry out the evaluation. Link to a screenshot/copy if possible. If there is a separate introductory interface/page, include it under Question 3.2.4.

\subsection*{\qsecbox{3.3.5: How free are evaluators regarding when and how quickly to carry out evaluations? Select all that apply. In all cases, provide details in the text box under `Other'.}}

\noindent\textit{Check-box options (select all that apply)}:  
\vspace{-.1cm}

\begin{enumerate}[itemsep=0cm,leftmargin=0.5cm,label={\small $\square$}]
    \item \egcvalue{evaluators have to complete each individual assessment within a set time}: evaluators are timed while carrying out each assessment and cannot complete the assessment once time has run out.
    \item \egcvalue{evaluators have to complete the whole evaluation in one sitting}: partial progress cannot be saved and the evaluation returned to on a later occasion.
    \item \egcvalue{neither of the above}: Choose this option if neither of the above are the case in the experiment.
    \item \egcvalue{Other (please specify)}: Use this space to describe any other way in which time taken or number of sessions used by evaluators is controlled in the experiment, and to provide additional details for any of the options selected above.
\end{enumerate}

\subsection*{\qsecbox{3.3.6: Are evaluators told they can ask questions about the evaluation and/or provide feedback? Select all that apply. In all cases, provide details in the text box under `Other'.}}

\noindent\textit{Check-box options (select all that apply)}:  
\vspace{-.1cm}

\begin{enumerate}[itemsep=0cm,leftmargin=0.5cm,label={\small $\square$}]
    \item \egcvalue{evaluators are told they can ask any questions during/after receiving initial training/instructions, and before the start of the evaluation}: evaluators are told explicitly that they can ask  questions about the evaluation experiment \textit{before} starting on their assessments, either during or after training.
    \item \egcvalue{evaluators are told they can ask any questions during the evaluation}: evaluators are told explicitly that they can ask  questions about the evaluation experiment \textit{during} their assessments.
    \item \egcvalue{evaluators are asked for feedback and/or comments after the evaluation, e.g.\ via an exit questionnaire or a comment box}: evaluators are explicitly asked to provide feedback and/or comments about the experiment \textit{after} their assessments, either verbally or in written form.
    \item \egcvalue{None of the above}: Choose this option if none of the above are the case in the experiment.
    \item \egcvalue{Other (please specify)}: use this space to describe any other ways you provide for evaluators to ask questions or provide feedback.
\end{enumerate}

\subsection*{\qsecbox{3.3.7: What are the experimental conditions in which evaluators carry out the evaluations? If none match, select `Other’ and describe.}}

\noindent\textit{Multiple-choice options (select one)}:  
\vspace{-.1cm}

\begin{enumerate}[itemsep=0cm,leftmargin=0.5cm,label={\LARGE $\circ$}]
    \item \egcvalue{evaluation carried out by evaluators at a place of their own choosing, e.g.\ online, using a paper form, etc.}: evaluators are given access to the tool or form specified in Question 3.3.2, and subsequently choose where to carry out their evaluations.
    \item \egcvalue{evaluation carried out in a lab, and conditions are the same for each evaluator}: evaluations are carried out in a lab, and conditions in which evaluations are carried out \textit{are} controlled to be the same, i.e.\ the different evaluators all carry out the evaluations in identical conditions of quietness, same type of computer, same room, etc. Note we're not after very fine-grained differences here, such as time of day  or temperature, but the line is difficult to draw, so some judgment is involved here.
    \item \egcvalue{evaluation carried out in a lab, and conditions vary for different evaluators}: choose this option if evaluations are carried out in a lab, but the preceding option does not apply, i.e.\ conditions in which evaluations are carried out are \textit{not} controlled to be the same.
    \item \egcvalue{evaluation carried out in a real-life situation, and conditions are the same for each evaluator}: evaluations are carried out in a real-life situation, i.e.\ one that would occur whether or not the evaluation was carried out (e.g.\ evaluating a dialogue system deployed in a live chat function on a website), and conditions in which evaluations are carried out \textit{are} controlled to be the same. 
    \item \egcvalue{evaluation carried out in a real-life situation, and conditions vary for different evaluators}: choose this option if evaluations are carried out in a real-life situation, but the preceding option does not apply, i.e.\ conditions in which evaluations are carried out are \textit{not} controlled to be the same.
    \item \egcvalue{evaluation carried out outside of the lab, in a situation designed to resemble a real-life situation, and conditions are the same for each evaluator}: evaluations are carried out outside of the lab, in a situation intentionally similar to a real-life situation (but not actually a real-life situation), e.g.\ user-testing a navigation system where the destination is part of the evaluation design, rather than chosen by the user. Conditions in which evaluations are carried out \textit{are} controlled to be the same. 
    \item \egcvalue{evaluation carried out outside of the lab, in a situation designed to resemble a real-life situation, and conditions vary for different evaluators}: choose this option if evaluations are carried out outside of the lab, in a situation intentionally similar to a real-life situation, but the preceding option does not apply, i.e.\ conditions in which evaluations are carried out are \textit{not} controlled to be the same. 
    \item \egcvalue{Other (please specify)}: Use this space to provide additional, or alternative, information about the conditions in which evaluators carry out assessments, not covered by the options above.
\end{enumerate}

\vspace{-.3cm}
\subsection*{\qsecbox{3.3.8:  Unless the evaluation is carried out at a place of the evaluators'  own choosing, briefly describe the (range of different) conditions in which evaluators carry out the evaluations.}}

\noindent\textit{What to enter in the text box}: use this space to describe the variations in the conditions in which evaluators carry out the evaluation, for both situations where those variations are controlled, and situations where they are not controlled.

\vspace{.3cm}

\section{Quality Criterion \textit{n} -- Definition and Operationalisation}
\label{sec:criteria}

Questions in this section collect information about the $n$th quality criterion assessed in the single human evaluation experiment that this sheet is being completed for. The HEDS 1.0 form allows this section to be completed repeatedly, for up to 10 different quality criteria (see further explanation at the end of the section).

For more information, in particular about quality criterion properties and evaluation mode properties, see \citet{belz-etal-2020-disentangling}.

\subsection{Quality criterion properties (Questions 4.1.1--4.1.3)}

Questions 4.1.1--4.1.3 capture the aspect of quality that is assessed by a given quality criterion in terms of three orthogonal properties. They help determine e.g.\ whether or not the same aspect of quality is being evaluated in different evaluation experiments. The three properties characterise quality criteria in terms of (i) what type of quality is being assessed; (ii) what aspect of the system output is being assessed; and (iii) whether system outputs are assessed in their own right or with reference to some system-internal or system-external frame of reference.

\vspace{-.3cm}
\subsection*{\qsecbox{Question 4.1.1:  What type of quality is assessed by the quality criterion?}}
\vspace{-.1cm}

\vspace{.3cm}
\noindent\textit{Multiple-choice options (select one)}:  
\vspace{-.1cm}

\begin{enumerate}[itemsep=0cm,leftmargin=0.5cm,label={\LARGE $\circ$}]
    \item \egcvalue{Correctness}: select this option if it is possible to state,  generally for all outputs,  the conditions under which outputs are maximally correct (hence of maximal quality).  E.g.\ for Grammaticality, outputs are (maximally) correct if they contain no grammatical errors; for Semantic Completeness, outputs are correct if they express all the content in the input.
    \item \egcvalue{Goodness}: select this option if, in contrast to correctness criteria, there is no single, general mechanism for deciding when outputs are maximally good, only for deciding for two outputs which is better and which is worse. E.g.\ for Fluency, even if outputs contain no disfluencies, there may be other ways in which any given output could be more fluent.
    \item \egcvalue{Features}: choose this option if, in terms of property $X$ captured by the criterion, outputs are not generally better if they are more $X$, but instead, depending on evaluation context, more $X$ may be better or less $X$ may be better. E.g.\ outputs can be more specific or less specific, but it’s not the case that outputs are, in the general case, better when they are more specific.
\end{enumerate}

\subsection*{\qsecbox{Question 4.1.2:  Which aspect of system outputs is assessed by the quality criterion?}}
\vspace{-.1cm}

\vspace{.3cm}
\noindent\textit{Multiple-choice options (select one)}:  
\vspace{-.1cm}

\begin{enumerate}[itemsep=0cm,leftmargin=0.5cm,label={\LARGE $\circ$}]
    \item \egcvalue{Form of output}: choose this option if the criterion assesses the form of outputs alone, e.g.\ Grammaticality is only about the form, a sentence can be grammatical yet be wrong or nonsensical in terms of content.
    \item \egcvalue{Content of output}: choose this option if the criterion assesses the content/meaning of the output alone, e.g.\ Meaning Preservation only assesses output content; two sentences can be considered to have the same meaning, but differ in form.
    \item \egcvalue{Both form and content of output}: choose this option if the criterion assesses outputs as a whole, not just form or just content. E.g.\ Coherence is a property of outputs as a whole, either form or meaning can detract from it.
\end{enumerate}

\subsection*{\qsecbox{Question 4.1.3: Is each output assessed for quality in its own right, or with reference to a system-internal or external frame of reference?}}

\noindent\textit{Multiple-choice options (select one)}:  
\vspace{-.1cm}

\begin{enumerate}[itemsep=0cm,leftmargin=0.5cm,label={\LARGE $\circ$}]
    \item \egcvalue{Quality of output in its own right}: choose this option if output quality is assessed without referring to anything other than the output itself, i.e.\ no  system-internal or external frame of reference. E.g.\ Poeticness is assessed by considering (just) the output and how poetic it is. 
    \item \egcvalue{Quality of output relative to the input}:  choose this option if output quality is assessed  relative to the input. E.g.\ Answerability is the degree to which the output question can be answered from information in the input.
    \item \egcvalue{Quality of output relative to a system-external frame of reference}: choose this option if output quality is assessed with reference to system-external information, such as a knowledge base, a person’s individual writing style, or the performance of an embedding system. E.g.\ Factual Accuracy assesses outputs relative to a source of real-world knowledge.
\end{enumerate}

\subsection{Evaluation mode properties (Questions 4.2.1--4.2.3)}

Questions 4.2.1--4.2.3 record properties that are orthogonal to quality criteria, i.e.\ any given quality criterion can in principle be combined with any of the modes (although some combinations are more common than others). 

\vspace{-.3cm}
\subsection*{\qsecbox{Question 4.2.1: Does an individual assessment involve an objective or a subjective judgment?}}

\noindent\textit{Multiple-choice options (select one)}:  
\vspace{-.1cm}

\begin{enumerate}[itemsep=0cm,leftmargin=0.5cm,label={\LARGE $\circ$}]
    \item \egcvalue{Objective}: Examples of objective assessment include any automatically counted or otherwise quantified measurements such as mouse-clicks, occurrences in text, etc. Repeated assessments of the same output with an objective-mode evaluation method always yield the same score/result.
    \item \egcvalue{Subjective}: Subjective assessments involve ratings, opinions and preferences by evaluators. Some criteria lend themselves more readily to subjective assessments, e.g.\ Friendliness of a conversational agent, but an objective measure e.g.\ based on lexical markers is also conceivable.
\end{enumerate}
    
\subsection*{\qsecbox{Question 4.2.2: Are outputs assessed in absolute or relative terms?}}
\vspace{-.1cm}

\vspace{.3cm}
\noindent\textit{Multiple-choice options (select one)}:  
\vspace{-.1cm}

\begin{enumerate}[itemsep=0cm,leftmargin=0.5cm,label={\LARGE $\circ$}]
    \item \egcvalue{Absolute}: choose this option if evaluators are shown outputs from a single system during each individual assessment. 
    \item \egcvalue{Relative}: choose this option if evaluators are shown outputs from multiple systems at the same time during assessments, typically ranking or preference-judging them.
\end{enumerate}

\vspace{-.3cm}
\subsection*{\qsecbox{Question 4.2.3: Is the evaluation intrinsic or extrinsic?}}

\noindent\textit{Multiple-choice options (select one)}:  
\vspace{-.1cm}

\begin{enumerate}[itemsep=0cm,leftmargin=0.5cm,label={\LARGE $\circ$}]
    \item \egcvalue{Intrinsic}: Choose this option if quality of outputs is assessed \textit{without} considering their \textit{effect} on something external to the system, e.g.\ the performance of an embedding system or of a user at a task.
    \item \egcvalue{Extrinsic}: Choose this option if quality of outputs is assessed in terms of their \textit{effect} on something external to the system such as the performance of an embedding system or of a user at a task.
\end{enumerate}

\subsection{Response elicitation (Questions 4.3.1--4.3.11)}

Questions 4.3.1--4.3.11 record information about how responses are elicited for the quality criterion this section is being completed for.

\vspace{-.3cm}
\subsection*{\qsecbox{Question 4.3.1: What do you call the quality criterion in explanations/interfaces to evaluators?  Enter `N/A' if criterion not named.}}

\noindent\textit{What to enter in the text box}: the name you use to refer to the quality criterion in explanations and/or interfaces created for evaluators. Examples of quality criterion names include Fluency, Clarity, Meaning Preservation. If no name is used, state `N/A'.

\subsection*{\qsecbox{Question 4.3.2:  What definition do you give for the quality criterion in explanations/interfaces to evaluators? Enter `N/A' if no definition given.}}

\noindent\textit{What to enter in the text box}: Copy and past the verbatim definition you give to evaluators to explain the quality criterion they're assessing. If you don't explicitly call it a definition, enter the nearest thing to a definition you give them.  If you don't give any definition, state `N/A'.

\subsection*{\qsecbox{Question 4.3.3:  Size of scale or other rating instrument (i.e.\ how many different possible values there are). Answer should be an integer or `continuous' (if it's not possible to state how many possible responses there are). Enter `N/A' if there is no rating instrument.}}

\noindent\textit{What to enter in the text box}: The number of different response values for this quality criterion. E.g.\ for a 5-point Likert scale, the size to enter is 5. For two-way forced-choice preference judgments, it is 2; if there's also a no-preference option, enter 3. For a slider that is mapped to 100 different values for the purpose of recording assessments, the size to enter is 100. If no rating instrument is used (e.g.\ when evaluation gathers post-edits or qualitative feedback only), enter `N/A'.

\subsection*{\qsecbox{Question 4.3.4: List or range of possible values of the scale or other rating instrument. Enter `N/A', if there is no rating instrument.}} 

\noindent\textit{What to enter in the text box}: list, or give the range of, the possible values  of the rating instrument. The list or range should be of the size specified in Question 4.3.3. If there are too many to list, use a range. E.g.\ for two-way forced-choice preference judgments, the list entered might be \textit{A better, B better}; if there's also a no-preference option, the list 
might be \textit{A better, B better, neither}. For a slider that is mapped to 100 different values for the purpose of recording assessments, the range \textit{1--100} might be entered. If no rating instrument is used (e.g.\ when evaluation gathers post-edits or qualitative feedback only), enter 'N/A'.

\subsection*{\qsecbox{Question 4.3.5:  How is the scale or other rating instrument presented to evaluators? If none match, select `Other’ and describe.}}

\noindent\textit{Multiple-choice options (select one)}:  
\vspace{-.1cm}

\begin{enumerate}[itemsep=0cm,leftmargin=0.5cm,label={\LARGE $\circ$}]
    \item \egcvalue{Multiple-choice options}: choose this option if evaluators select exactly one of multiple options.
    \item \egcvalue{Check-boxes}: choose this option if evaluators select any number of options from multiple given options.
    \item \egcvalue{Slider}: choose this option if evaluators move a pointer on a slider scale to the position corresponding to their assessment.
    \item \egcvalue{N/A (there is no rating instrument)}: choose this option if there is no rating instrument.
    \item \egcvalue{Other (please specify)}: choose this option if there is a rating instrument, but none of the above adequately describe the way you present it to evaluators. Use the text box to describe the rating instrument and link to a screenshot.
\end{enumerate}

\subsection*{\qsecbox{Question 4.3.6:  If there is no rating instrument, describe briefly what task the evaluators perform (e.g.\ ranking multiple outputs, finding information, playing a game, etc.), and what information is recorded. Enter `N/A' if there is a rating instrument.}}

\noindent\textit{What to enter in the text box}:  If (and only if) there is no rating instrument, i.e.\ you entered `N/A' for Questions 4.3.3--4.3.5, describe the task evaluators perform in this space. Otherwise, here enter `N/A' if there \textit{is} a rating instrument.

\subsection*{\qsecbox{Question 4.3.7:  What is the verbatim question, prompt or instruction given to evaluators (visible to them during each individual assessment)?  }}

\noindent\textit{What to enter in the text box}:  Copy and paste the verbatim text that evaluators see during each assessment, that is intended to convey the evaluation task to them. E.g.\ \textit{Which of these texts do you prefer?} Or \textit{Make any corrections to this text that you think are necessary in order to improve it to the point where you would be happy to provide it to a client.} 

\subsection*{\qsecbox{Question 4.3.8:  Form of response elicitation. If none match, select `Other' and describe.}}

\noindent\textit{Multiple-choice options (select one)}:\footnote{Explanations adapted from \citet{howcroft-etal-2020-twenty}.}
\vspace{-.1cm}

\begin{enumerate}[itemsep=0cm,leftmargin=0.5cm,label={\LARGE $\circ$}]
    \item \egcvalue{(dis)agreement with quality statement}: Participants specify the degree to which they agree with a given quality statement by indicating their agreement on a rating instrument. The rating instrument is labelled with degrees of agreement and can additionally have numerical labels.  E.g.\ \textit{This text is fluent --- 1=strongly disagree...5=strongly agree}.
    \item \egcvalue{direct quality estimation}: Participants are asked to provide a rating using a rating instrument, which typically (but not always) mentions the quality criterion explicitly. E.g.\ \textit{How fluent is this text? --- 1=not at all fluent...5=very fluent}.
    \item \egcvalue{relative quality estimation (including ranking)}: Participants evaluate two or more items in terms of which is better.
    E.g.\ \textit{Rank these texts in terms of fluency}; \textit{Which of these texts is more fluent?}; \textit{Which of these items do you prefer?}.
    \item \egcvalue{counting occurrences in text}: Evaluators are asked to count how many times some type of phenomenon occurs, e.g.\ the number of facts contained in the output that are inconsistent with the input.
    \item \egcvalue{qualitative feedback (e.g.\ via comments entered in a text box)}: Typically, these are responses to open-ended questions in a survey or interview.
    \item \egcvalue{evaluation through post-editing/annotation}: Choose this option if the evaluators' task consists of editing or inserting annotations in text. E.g.\ evaluators may perform error correction and edits are then automatically measured to yield a numerical score.
    \item \egcvalue{output classification or labelling}: Choose this option if evaluators assign outputs to categories. E.g.\ \textit{What is the overall sentiment of this piece of text? --- Positive/neutral/negative.}
    \item \egcvalue{user-text interaction measurements}: choose this option if participants in the evaluation experiment interact with a text in some way, and measurements are taken of their interaction. E.g.\ reading speed, eye movement tracking, comprehension questions, etc. Excludes situations where participants are given a task to solve and their performance is measured which comes under the next option.
    \item \egcvalue{task performance measurements}: choose this option if participants in the evaluation experiment are given a task to perform, and measurements are taken of their performance at the task.  E.g.\ task is finding information, and task performance measurement is task completion speed and success rate.
    \item \egcvalue{user-system interaction measurements}: choose this option if participants in the evaluation experiment interact with a system in some way, while measurements are taken of their interaction. E.g.\ duration of interaction, hyperlinks followed, number of likes, or completed sales.
    \item \egcvalue{Other (please specify)}: Use the text box to describe the form of response elicitation used in assessing the quality criterion if it doesn't fall in any of the above categories.
\end{enumerate}

\subsection*{\qsecbox{Question 4.3.9:  How are raw responses from participants aggregated or otherwise processed to obtain reported scores for this quality criterion? State if no scores reported.}}
\vspace{-.1cm}

\vspace{.3cm}
\noindent\textit{What to enter in the text box}:  normally a set of separate assessments is collected from evaluators and is converted to the results as reported. Describe here the method(s) used in the conversion(s). E.g.\ macro-averages or micro-averages are computed from numerical scores to provide summary, per-system results.

\vspace{-.3cm}
\subsection*{\qsecbox{Question 4.3.10:  Method(s) used for determining effect size and significance of findings for this quality criterion.}}
\vspace{-.1cm}

\vspace{.3cm}
\noindent\textit{What to enter in the text box}: A list of  methods used for calculating the effect size and significance of any results, both as reported in the paper given in Question 1.1, for this quality criterion. If none calculated, state `None'.

\vspace{-.3cm}
\subsection*{\qsecbox{Question 4.3.11:  Has the inter-annotator and intra-annotator agreement between evaluators for this quality criterion been measured? If yes, what method was used, and what are the agreement scores?}}

\noindent\textit{What to enter in the text box}: the methods used to compute, and results obtained from, any measures of inter-annotator and intra-annotator agreement obtained for the quality criterion.

\vspace{.3cm}
\noindent The section ends with the question \textbf{Is there another quality criterion in the evaluation experiment that you haven't completed this section for yet?} If \textbf{Yes} is selected, another copy of the quality criterion section opens up, to be completed for the next criterion. If \textbf{No}, the next section will be the Ethics section below. The maximum number of criteria the form can currently be completed for is 10. We can add more on request.

\section{Ethics}\label{sec:ethics}

The questions in this section relate to ethical aspects of the evaluation. Information can be entered in the text box provided, and/or by linking to a source where complete information can be found.

\vspace{-.3cm}
\subsection*{\qsecbox{Question 5.1: Has the evaluation experiment this sheet is being completed for, or the larger study it is part of, been approved by a research ethics committee? If yes, which research ethics committee?}}
\vspace{-.1cm}

\vspace{.3cm}
\noindent\textit{What to enter in the text box}: Typically, research organisations, universities and other higher-education institutions require some form ethical approval before experiments involving human participants, however innocuous, are permitted to proceed. Please provide here the name of the body that approved the experiment, or state `No' if approval has not (yet) been obtained.

\vspace{-.3cm}
\subsection*{\qsecbox{Question 5.2: Do any of the system outputs (or human-authored stand-ins) evaluated, or do any of the responses collected, in the experiment contain personal data (as defined in GDPR Art. 4, §1: https://gdpr.eu/article-4-definitions/)? If yes, describe data and state how addressed.}}
\vspace{-.1cm}

\vspace{.3cm}
\noindent\textit{What to enter in the text box}: State `No' if no personal data as defined by GDPR was recorded or collected, otherwise explain how conformity with GDPR requirements such as privacy and security was ensured, e.g.\ by linking to the (successful) application for ethics approval from Question 5.1.

\vspace{-.3cm}
\subsection*{\qsecbox{Question 5.3: Do any of the system outputs (or human-authored stand-ins) evaluated, or do any of the responses collected, in the experiment contain special category information (as defined in GDPR Art. 9, §1: https://gdpr.eu/article-9-processing-special-categories-of-personal-data-prohibited/)? If yes, describe data and state how addressed.}}
\vspace{-.1cm}

\vspace{.3cm}
\noindent\textit{What to enter in the text box}: State `No' if no special-category data as defined by GDPR was recorded or collected, otherwise explain how conformity with GDPR requirements relating to special-category data was ensured, e.g.\ by linking to the (successful) application for ethics approval from Question 5.1.

\vspace{-.3cm}
\subsection*{\qsecbox{Question 5.4:  Have any impact assessments been carried out for the evaluation experiment, and/or any data collected/evaluated in connection with it? If yes, summarise approach(es) and outcomes.}}

\noindent\textit{What to enter in the text box}: Use this box to describe any \textit{ex ante} or \textit{ex post} impact assessments that have been carried out in relation to the evaluation experiment, such that the assessment plan and process, as well as the outcomes, were captured in written form. Link to documents if possible. Types of impact assessment include data protection impact assessments, e.g.\ under GDPR.\footnote{\footnotesize \url{https://ico.org.uk/for-organisations/guide-to-data-protection/guide-to-the-general-data-protection-regulation-gdpr/accountability-and-governance/data-protection-impact-assessments/}}Environmental and social impact assessment frameworks are also available.

\section{Future Work}

We welcome feedback on any aspect of the Human Evaluation Datasheet. We envisage the template and questions to evolve over time partly in response to such feedback. Currently planned work includes (i) increasing the generality of the template to make it equally suitable for all forms of human evaluation in NLP, and (ii) creating a repository of completed datasheets.

\bibliography{main}
\bibliographystyle{acl_natbib}

\end{document}